\pdfoutput=1

\documentclass[11pt]{article}

\usepackage{EMNLP2022}

\usepackage{times}
\usepackage{latexsym}

\usepackage[T1]{fontenc}

\usepackage[utf8]{inputenc}

\usepackage{microtype}

\usepackage{graphicx}
\usepackage{multirow}
\usepackage{amsmath}
\usepackage{amsfonts}
\usepackage{bm}
\usepackage{booktabs}
\usepackage{subfigure}
\usepackage{verbatim}
\usepackage{amssymb}
\usepackage{enumitem}

\usepackage{inconsolata}

\title{Improving Aspect Sentiment Quad Prediction via Template-Order \\ Data Augmentation}

\author{Mengting Hu\textsuperscript{1} \quad Yike Wu\textsuperscript{2} \quad Hang Gao\textsuperscript{3}\thanks{\; Corresponding author.} \quad Yinhao Bai\textsuperscript{1} \quad {\bf Shiwan Zhao\textsuperscript{}\thanks{\; Independent researcher.}} \\
\textsuperscript{1} College of Software, Nankai University \\
\textsuperscript{2} School of Journalism and Communication, Nankai University \\
\textsuperscript{3} Institute for Public Safety Research, Tsinghua University \\
{\tt \{mthu, wuyike\}@nankai.edu.cn,} {\tt gaohang@mail.tsinghua.edu.cn} \\ {\tt yinhao@mail.nankai.edu.cn,} 
{\tt zhaosw@gmail.com}
}

\begin{document}
\maketitle
\begin{abstract}

Recently, aspect sentiment quad prediction (ASQP) has become a popular task in the field of aspect-level sentiment analysis. Previous work utilizes a predefined template to paraphrase the original sentence into a structure target sequence, which can be easily decoded as quadruplets of the form (\emph{aspect category}, \emph{aspect term}, \emph{opinion term}, \emph{sentiment polarity}). The template involves the four elements in a fixed order. However, we observe that this solution contradicts with the order-free property of the ASQP task, since there is no need to fix the template order as long as the quadruplet is extracted correctly. Inspired by the observation, we study the effects of template orders and find that some orders help the generative model achieve better performance. It is hypothesized that different orders provide various views of the quadruplet. Therefore, we propose a simple but effective method to identify the most proper orders, and further combine multiple proper templates as data augmentation to improve the ASQP task. Specifically, we use the pre-trained language model to select the orders with minimal entropy. By fine-tuning the pre-trained language model with these template orders, our approach improves the performance of quad prediction, and outperforms state-of-the-art methods significantly in low-resource settings\footnote{Experimental codes and data are available at: \url{https://github.com/hmt2014/AspectQuad}.}.
\end{abstract}


\section{Introduction}
The aspect sentiment quad prediction (ASQP) task, aiming to extract aspect quadruplets from a review sentence, becomes popular recently \cite{zhang-etal-2021-aspect-sentiment,cai2021aspect}. The quadruplet consists of four sentiment elements: 1) \emph{aspect category (ac)} indicating the aspect class; 2) \emph{aspect term (at)} which is the specific aspect description; 3) \emph{opinion term (ot)} which is the opinion expression towards the aspect; 4) \emph{sentiment polarity (sp)} denoting the sentiment class of the aspect. For example, the sentence \emph{``The service is good and the restaurant is clean.''} contains two quadruplets (\emph{service general}, \emph{service}, \emph{good}, \emph{positive}) and (\emph{ambience general}, \emph{restaurant}, \emph{clean}, \emph{positive}).

\begin{figure}[t]
\centering
\includegraphics[width=0.48\textwidth]{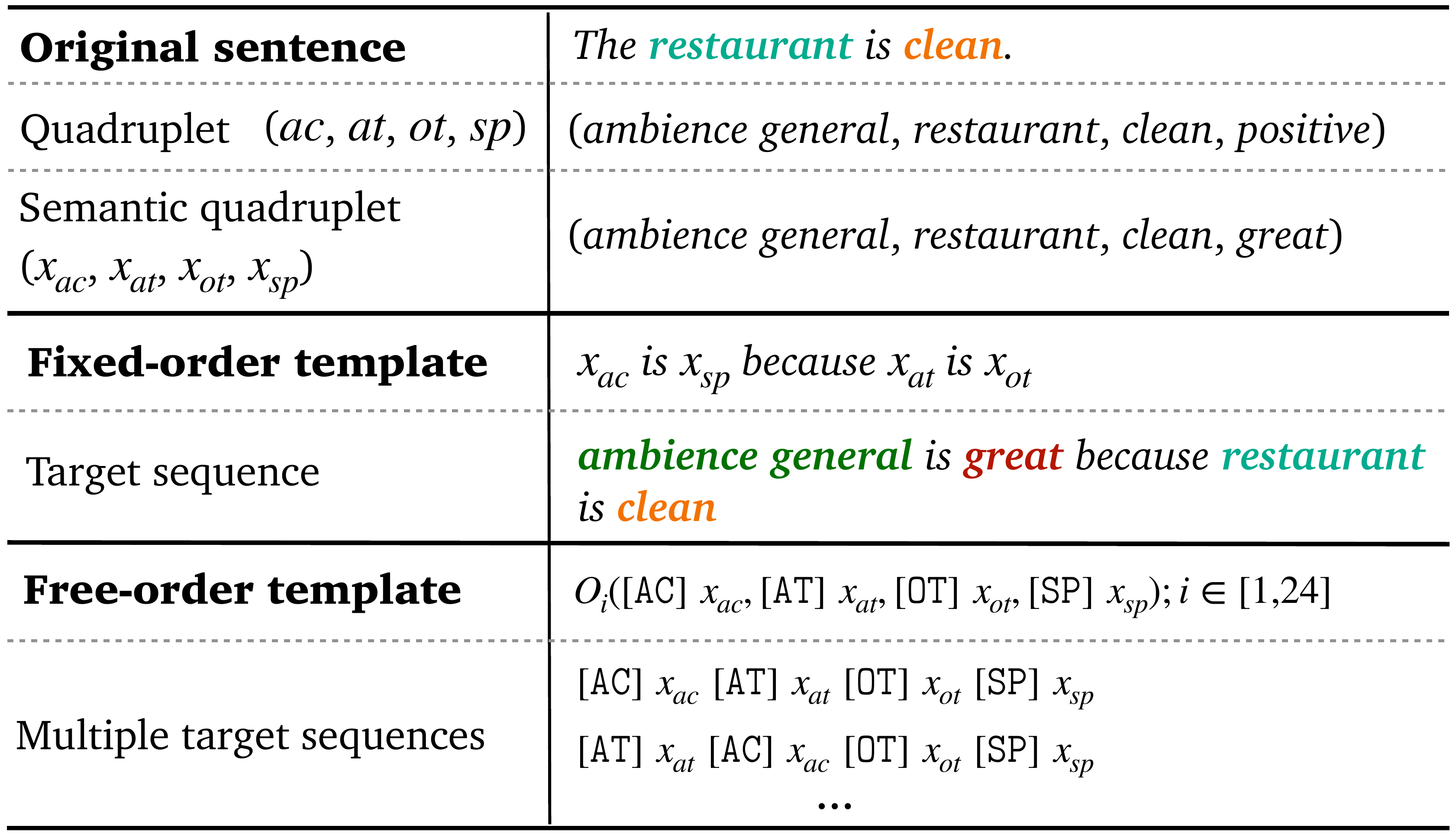} 
\caption{An example sentence is paraphrased into a target sequence with a fixed-order template \cite{zhang-etal-2021-aspect-sentiment}. Our approach employs special markers to form free-order templates and generates multiple target sequences. $O_i$ is the $i$-th order permutation of the four elements.}
\label{fig:examples}
\end{figure}

To extract aspect sentiment quadruplets, \citet{zhang-etal-2021-aspect-sentiment} propose a new paradigm which transforms the quadruplet extraction into paraphrase generation problem. With pre-defined rules, they first map the four elements of ($ac$, $at$, $ot$, $sp$) into semantic values ($x_{ac}$, $x_{at}$, $x_{ot}$, $x_{sp}$), which are then fed into a template to obtain a nature language target sequence. As shown in Figure \ref{fig:examples}, the original sentence is \emph{``re-writen''} into a target sequence by paraphrasing. After fine-tuning the pre-trained language model \cite{JMLR:v21:20-074} in such a sequence-to-sequence learning manner, the quadruplets can be disentangled from the target sequence.

Though promising is this paradigm, one issue is that the decoder of the generative pre-trained language model \cite{JMLR:v21:20-074} is unidirectional \cite{vinyals2015order}, which outputs the target sequence from the beginning of the sequence to its end. Thus four elements of a quadruplet are modeled in a fixed order $\{x_{ac}\rightarrow{x_{sp}}\rightarrow{x_{at}}\rightarrow{x_{ot}}\}$. Yet ASQP is not a typical generation task. 
There is no need to fix the element order of the quadruplet as long as it can be extracted accurately. Aspect sentiment quadruplet has the order-free property, suggesting that various orders, such as $\{x_{ac}\rightarrow{x_{sp}}\rightarrow{x_{at}}\rightarrow{x_{ot}}\}$ and $\{x_{at}\rightarrow{x_{ac}}\rightarrow{x_{ot}}\rightarrow{x_{sp}}\}$, are all correct.

In light of this observation, our curiosity is triggered: Does the order of the four elements impact the generative pre-trained language models' performances? Thus we conduct a pilot experiment. The four elements are concatenated with commas, thus we could switch their orders in a flexible manner and obtain order permutations. It is found that some template orders can help the generative model perform better. Even only concatenating with commas, some orders outperform the state-of-the-art.  %





It is hypothesized that different orders provide various views of the quadruplet. Therefore, we propose a simple but effective method to identify the most proper orders, and further combine multiple proper templates as data augmentation to improve the ASQP task. Concretely, we use the pre-trained language model \cite{JMLR:v21:20-074} to select the orders with minimal entropy. Such template orders can better promote the potential of the pre-trained language model. To jointly fine-tune these template orders together, inspired by \citet{paolini2021structured}, we design special markers for the four elements, respectively. The markers help to disentangle quadruplets by recognizing both the types and their values of the four elements from the target sequence. In this way, the template orders do not need to be fixed in advance.




In summary, the contributions of this work are three-fold:
\begin{itemize}
    \item We study the effects of template orders in the ASQP task, showing that some orders perform better. To the best of our knowledge, this work is the first attempt to investigate ASQP from the template order perspective.
    \item We propose to select proper template orders by minimal entropy computed with pre-trained language models. The selected orders are roughly consistent with their ground-truth performances. 
    \item Based on the order-free property of the quadruplet, we further combine multiple proper templates as data augmentation to improve the ASQP task. 
    Experimental results demonstrate that our approach outperforms state-of-the-art methods and has significant gains in low-resource settings.
\end{itemize}



\section{Preliminaries on Generative ASQP}
\label{sec:paraphrase}
\subsection{Paraphrase Generation}
Given a sentence $\bm{x}$, aspect sentiment quad prediction (ASQP) aims to extract all aspect-level quadruplets $\{({ac},{at},{ot},{sp})\}$. Recent paradigm for ASQP \cite{zhang-etal-2021-aspect-sentiment} formulates this task as a paraphrase generation problem. They first define projection functions to map quadruplet $({ac},{sp},{at},{ot})$ into semantic values $(x_{ac},x_{sp},x_{at},x_{ot})$. Concretely, 1) aspect category $ac$ is transformed into words, such as $x_{ac}=$\emph{``service general''} for $ac=$\emph{``service\#general''}; 2) if aspect term $at$ is explicit, $x_{at}={at}$, otherwise $x_{at}=$``it''; 3) if opinion term $ot$ are explicitly mentioned, $x_{ot}=ot$, otherwise it is mapped as \emph{``NULL''} if being implicitly expressed; 4) the sentiment polarity $sp\in$ \{\emph{positive}, \emph{neutral}, \emph{negative}\}, is mapped into words with sentiment semantics \{\emph{great}, \emph{ok}, \emph{bad}\}, respectively.

With the above rules, the values can better exploit the semantic knowledge from pre-trained language model. Then the values of quadruplet are fed into the template, which follows the cause and effect semantic relationship.
\begin{equation}
    x_{ac} \text{ is } x_{sp} \text{ because } x_{at} \text{ is } x_{ot}.
    \label{eq:paraphrase_template}
\end{equation}

It is worth noting that if a sentence describes multiple quadruplets, the paraphrases are concatenated with a special marker $\mathtt{[SSEP]}$ to obtain the final target sequence $\bm{y}$.

\subsection{Sequence-to-Sequence Learning}
The purpose of paraphrasing is consistent with the typical sequence-to-sequence problem. The encoder-decoder model is leveraged to \emph{``re-write''} the original sentence $\bm{x}$ into the target sequence $\bm{y}$. Assume the parameter is $\theta$, the overall objective is to model the conditional probability $p_\theta(\bm{y}|\bm{x})$. Specifically, at the $t$-th time step, the decoder output $\bm{y}_t$ is calculated with the input $\bm{x}$ and the previous outputs $\bm{y}_{<t}$, formulating as below.
\begin{equation}
    p_\theta(\bm{y}_{t+1}|\bm{x},\bm{y}_{<t+1})=\mathrm{softmax}(W^{\mathrm{T}}\bm{y}_t)
    \label{eq:conditional}
\end{equation}
where $W$ maps $\bm{y}_t$ into a vector, which can represent the probability distribution over the whole vocabulary set.


During training, a pre-trained encoder-decoder model, i.e. T5 \cite{JMLR:v21:20-074}, is chosen to initialize the parameter $\theta$ and fine-tuned with minimizing the cross-entropy loss.
\begin{equation}
    \mathcal{L}(\bm{x},\bm{y})=-\sum_{t=1}^n\mathrm{log}p_\theta(\bm{y}_t|\bm{x},\bm{y}_{<t})
\end{equation}
where $n$ is the length of the target sequence $\bm{y}$.

\begin{table}[]
\small
    \centering
    \setlength{\tabcolsep}{2.8mm}{
    \begin{tabular}{c|ccc}
    \toprule
    \multirow{2}{*}{Target Sequence} & \multicolumn{3}{c}{$\mathtt{Rest15}$} \\ 
    & $\mathtt{Pre}$ & $\mathtt{Rec}$ & $\mathtt{F1}$  \\
    \midrule
$x_{sp}$, $x_{ac}$, $x_{at}$, $x_{ot}$ & 45.55 & 46.34 & 45.94 \\
$x_{sp}$, $x_{ot}$, $x_{at}$, $x_{ac}$ & 46.12 & 47.52 & 46.81 \\
$x_{ac}$, $x_{ot}$, $x_{at}$, $x_{sp}$ & 47.07 & 47.85 & 47.46 \\
$x_{ac}$, $x_{sp}$, $x_{ot}$, $x_{at}$ & \textbf{47.60} & \textbf{48.75} & \textbf{48.17} \\
\toprule
\toprule
\multirow{2}{*}{Target Sequence} &  \multicolumn{3}{c}{$\mathtt{Rest16}$} \\
& $\mathtt{Pre}$ & $\mathtt{Rec}$ & $\mathtt{F1}$  \\
\midrule
$x_{ot}$, $x_{at}$, $x_{sp}$, $x_{ac}$ & 56.04 & 58.17 & 57.09 \\
$x_{ac}$, $x_{ot}$, $x_{sp}$, $x_{at}$ & 57.14 & 58.72 & 57.92 \\
$x_{at}$, $x_{sp}$, $x_{ac}$, $x_{ot}$ & 57.35 & 59.60 & 58.45 \\
$x_{ac}$, $x_{sp}$, $x_{at}$, $x_{ot}$ & \textbf{58.11} & \textbf{60.33} & \textbf{59.20} \\
\bottomrule
    \end{tabular}}
    \caption{Evaluation results on various template orders in terms of precision ($\mathtt{Pre}$, \%), recall ($\mathtt{Rec}$, \%) and F1 ($\mathtt{F1}$, \%) scores. All the reported results are the average of five runs. Full results are shown in the appendix, where some template orders outperform Paraphrase.}
    \label{table:result_pre}
\end{table}

\section{A Pilot Experiment}
As Eq. (\ref{eq:paraphrase_template}) displayed, this template forms a fixed order of four elements. Our curiosity is whether the quadruplet's order affects the performance of sequence-to-sequence learning. Therefore, we conduct a pre-experiment. By only concatenating with \emph{commas}, four elements can also be transformed into a target sequence. The orders can be switched in a more flexible way, compared with Eq. (\ref{eq:paraphrase_template}). There will be $4!=24$ permutations. During inference, quadruplets can be recovered by splitting them with commas. Based on the pre-experimental results, we have the following observations.


\textbf{Template order affects the performances of sequence-to-sequence learning.} \; Part of the experimental results on two datasets, i.e. $\mathtt{Rest15}$ and $\mathtt{Rest16}$, are shown in Table \ref{table:result_pre}. It is observed that on $\mathtt{Rest15}$, the $\mathtt{F1}$ score ranges from 45.94\% to 48.17\%. Similarly, on $\mathtt{Rest16}$, the $\mathtt{F1}$ score ranges from 57.09\% to 59.20\%. We draw an empirical conclusion that the template order also matters for the ASQP task. Moreover, a template has various performances on different datasets, which is hard to say some order is absolutely good.

\begin{figure}[t]
\centering
\subfigure[Results on $\mathtt{Rest15}$]{
\includegraphics[width=0.23\textwidth]{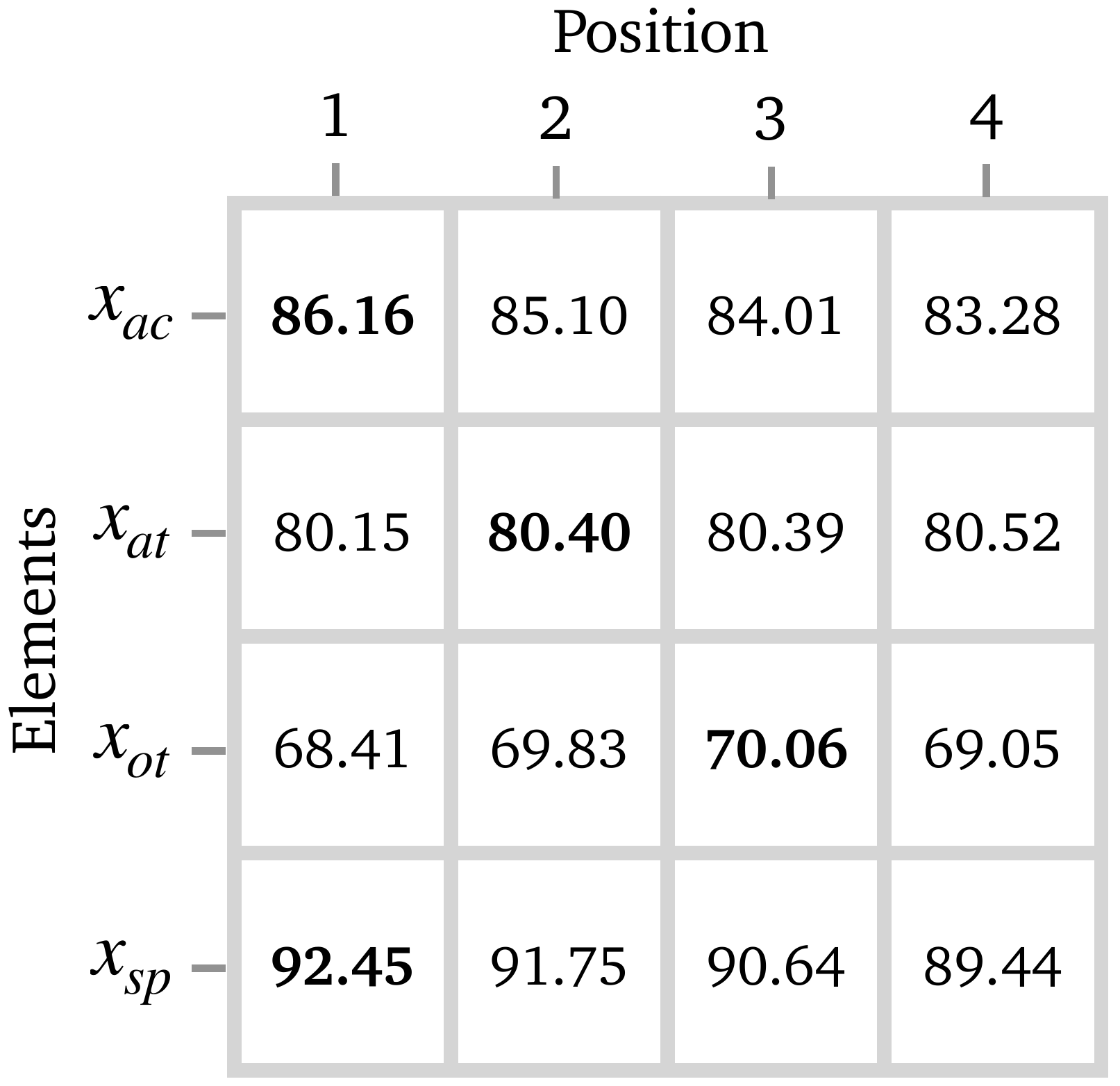}}
\subfigure[Results on $\mathtt{Rest16}$]{
\includegraphics[width=0.23\textwidth]{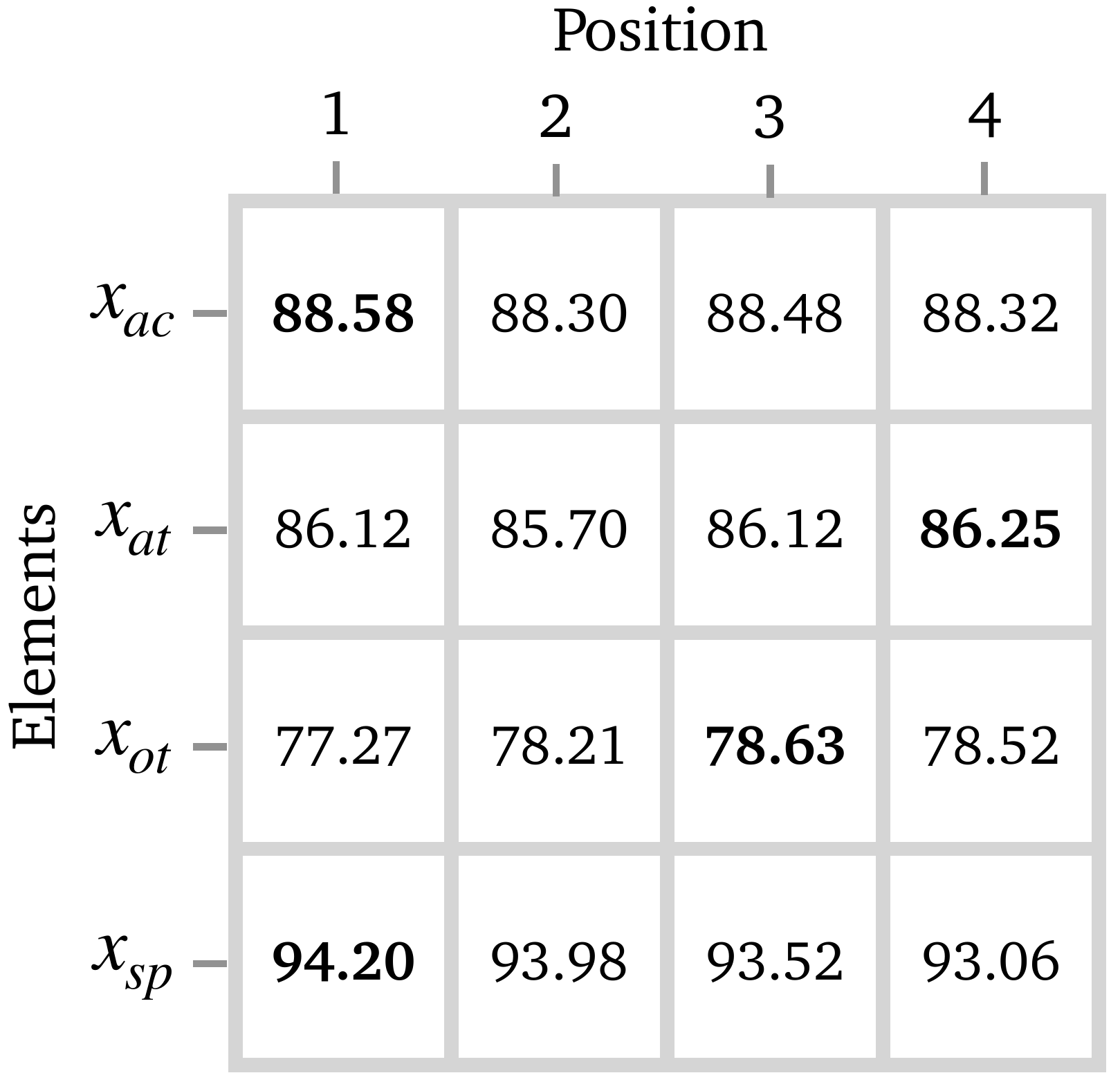}}
\caption{Evaluation results of each element at each position in terms of $\mathtt{F1}$ score (\%). The performances of different elements show different trends with the position. The best score of each row is marked in bold.}
\label{fig:figure_pre}
\end{figure}

\textbf{The performances of each element are connected to its position.} We further investigate the $\mathtt{F1}$ scores on each of the four elements. Given the $24$ permutations, there are $6$ templates for each element at each position. For example, there are $6$ templates $\{(., ., x_{ac}, .)\}$ of $x_{ac}$ at position $3$. 
In Figure \ref{fig:figure_pre}, we show the average $\mathtt{F1}$ scores of the six templates of each element at each position. 
We can see that the performances of the four elements have different trends with the positions. The $\mathtt{F1}$ scores of $x_{ac}$ and $x_{sp}$ both degrade when they are gradually placed backwards. Compared with the other three elements, $x_{at}$ is more stable on different positions while $x_{ot}$ has the worst performance in the first position. In addition to positions, it can also be observed that $x_{ac}$ and $x_{sp}$ achieve higher $\mathtt{F1}$ scores compared with $x_{at}$ and $x_{ot}$, showing various difficult extents of four elements. 



\section{Methodology}

As analyzed in the previous section, the template order influences the performances of both the quadruplet and its four elements. It is hypothesized that different orders provide various views of the quadruplet. We argue that combining multiple template orders may improve the ASQP task via data augmentation. However, using all $24$ permutations significantly increases the training time, which is inefficient. Therefore, we propose a simple method to select proper template orders by leveraging the nature of the pre-trained language model (i.e. T5). Then for the ASQP task, these selected orders are utilized to construct the target sequence $\bm{y}$ to fine-tune the T5 model. 

\begin{figure}[t]
\centering
\includegraphics[width=0.48\textwidth]{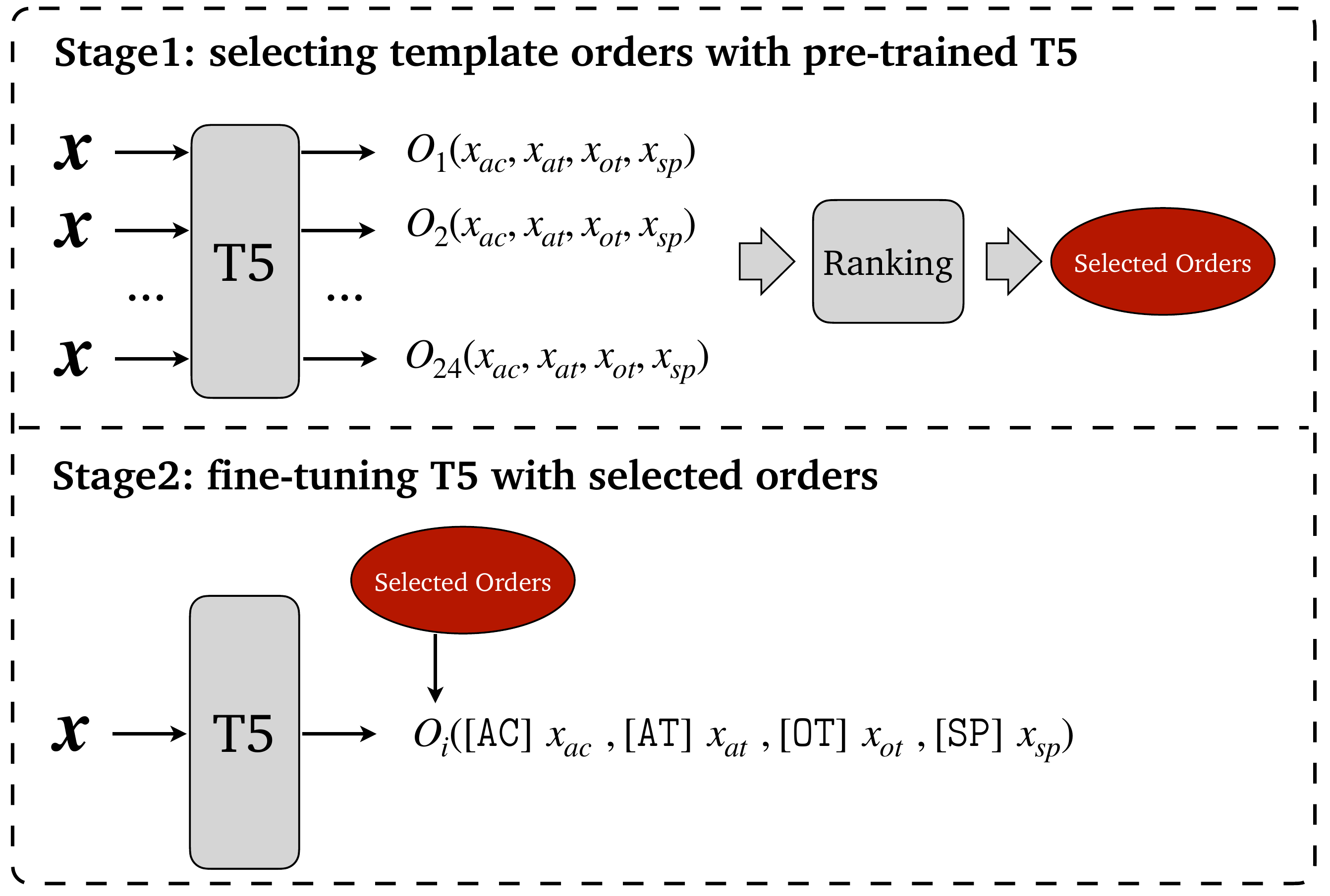} 
\caption{The proposed method is composed of two stages. The first stage aims to select template orders via pre-trained T5. The second stage constructs training samples with the selected orders and fine-tunes T5.}
\label{fig:model}
\end{figure}

Specifically, given an input sentence $\bm{x}$ and its quadruplets $\{({ac},{at},{ot},{sp})\}$, following \citet{zhang-etal-2021-aspect-sentiment}, we map them into semantic values $\{(x_{ac},x_{at},x_{ot},x_{sp})\}$. As shown in Figure \ref{fig:model}, our approach is composed of two stages, next which will be introduced in detail. 

\subsection{Selecting Template Orders}
Inspired by \cite{yuan2021bartscore,lu2021fantastically}, we choose template orders by evaluating them with the pre-trained T5. As shown in Figure \ref{fig:model}, given an input $\bm{x}$ and it quadruplets, we construct all 24 target sequences with multiple order mapping functions $O_i$, where $i\in{[1,24]}$. An example $O_i$ is shown below.
\begin{equation}
    O_i(x_{ac}, x_{at}, x_{ot}, x_{sp})=x_{at}\ x_{ac}\ x_{ot}\ x_{sp}
\end{equation} 
where the four values are concatenated with a simple space, without any other tokens such as commas, in a specific order $O_i$. In this way we can reduce the impact of noisy tokens, but focus more on the order. We also introduce the special symbol $\mathtt{[SSEP]}$ if there are multiple quadruplets in a sentence. Given multiple template orders, multiple target sequences $\bm{y}_{o_i}$ are constructed for an input $\bm{x}$.


Then we evaluate these target sequences with the entropy computed by the pre-trained T5. Here $\bm{y}_{o_i}$ is also fed into the decoder as teacher forcing \cite{williams1989learning}. The output logits $p_\theta$ of the decoder are utilized to compute the entropy.
\begin{equation}
    \mathcal{E}(\bm{y}_{o_i}|\bm{x})=-\frac{1}{n}\sum_{n}\sum_{|V|}{p_\theta}\mathrm{log} p_\theta
    \label{eq:entropy}
\end{equation}
where $n$ is length of target sequence and and $|V|$ is the size of the vocabulary set. 

Given the whole training set $\mathcal{T}$, we have $(\bm{x},\{\bm{y}_{o_i}\}_{i=1}^{24})$ for each instance by constructing template orders. Specifically, we design the following two template selection strategies. 


\vspace{6pt}
\noindent
\textbf{Dataset-Level Order (DLO)} \; To choose the dataset-level orders, we compute a score for each order on the whole training set.
\begin{equation}
    \mathcal{S}_{o_i}=\frac{1}{|\mathcal{T}|}\sum_{\mathcal{T}}\mathcal{E}(\bm{y}_{o_i}|\bm{x})
\end{equation}
where $\mathcal{S}_{o_i}$ denotes the average entropy of all instances for the template order $O_i$. Then by ranking these scores, template orders with smaller values are chosen. 

\vspace{6pt}
\noindent
\textbf{Instance-Level Order (ILO)} \; Different instances have various contexts and semantics, and tend to have their own proper template orders. Therefore, we also design to choose orders at the instance level. Similarly, the template orders of each instance with small values are chosen based on Eq. (\ref{eq:entropy}).


\subsection{Fine-tuning with Selected Orders}
Multiple template orders provide various views of a quadruplet. However, to train them jointly, an issue arises. If the four values are concatenated with a comma or only a blank space, the value type could not be identified during the inference. For example, when the output sequence \emph{``food quality, pasta, delicious, great''} is recovered to a quadruplet, the machine does not know the element types. Therefore, to deal with this issue, we design special markers to represent the structure of the information \cite{paolini2021structured}. The markers for $x_{ac}$, $x_{at}$, $x_{ot}$, $x_{sp}$ are $\mathtt{[AC]}$, $\mathtt{[AT]}$, $\mathtt{[OT]}$, $\mathtt{[SP]}$, respectively. Given an order, the target sequence is constructed:
\begin{equation*}
\begin{split}
    \bm{y}_{o_i}&=O_i(\mathtt{[AC]}\ x_{ac}, \mathtt{[AT]\ x_{at}}, \mathtt{[OT]}\ x_{ot}, \mathtt{[SP]}\ x_{sp}) \\
    &=\mathtt{[AT]}\ x_{at}\  \mathtt{[AC]}\ x_{ac}\  \mathtt{[SP]}\ x_{sp}\  \mathtt{[OT]}\ x_{ot}
\end{split}
\end{equation*}

Now we can train multiple orders together, meanwhile the quadruplet can be recovered by these special markers during the inferences. Note that previous data augmentation methods usually design multiple inputs for one label, such as word deletion and replacement \cite{gao2021simcse}, obtaining multiple input sentences. While our method constructs multiple labels for one input sequence. This is beneficial from the ASQP task's property when using generation-based models. 

\begin{table*}[]
\small
    \centering
    \begin{tabular}{l|ccc|ccc}
    \toprule
    \multirow{2}{*}{Methods} & \multicolumn{3}{c|}{$\mathtt{Rest15}$} & \multicolumn{3}{c}{$\mathtt{Rest16}$} \\
    & $\mathtt{Pre}$ & $\mathtt{Rec}$ & $\mathtt{F1}$ & $\mathtt{Pre}$ & $\mathtt{Rec}$ & $\mathtt{F1}$ \\
    \midrule
    HGCN-BERT+BERT-Linear$^*$ \cite{cai2020aspect} & 24.43 & 20.25 & 22.15 & 25.36 & 24.03 & 24.68 \\
    HGCN-BERT+BERT-TFM$^*$ \cite{cai2020aspect} & 25.55 & 22.01 & 23.65 & 27.40 & 26.41 & 26.90 \\
    TASO-BERT-Linear$^*$ \cite{wan2020target} & 41.86 & 26.50 & 32.46 & 49.73 & 40.70 & 44.77 \\
    TASO-BERT-CRF$^*$ \cite{wan2020target} & 44.24 & 28.66 & 34.78 & 48.65 & 39.68 & 43.71 \\
    Extract-Classify-ACOS \cite{cai2021aspect} & 35.64 & 37.25 & 36.42 & 38.40 & 50.93 & 43.77 \\
    GAS$^*$ \cite{zhang-etal-2021-towards-generative} & 45.31 & 46.70 & 45.98 & 54.54 & 57.62 & 56.04 \\
    Paraphrase$^*$ \cite{zhang-etal-2021-aspect-sentiment} & 46.16 & 47.72 & 46.93 & 56.63 & 59.30 & 57.93 \\
    \midrule
    DLO & 47.08 & 49.33 & 48.18 & \textbf{57.92} & \textbf{61.80} & \textbf{59.79} \\
    ILO & \textbf{47.78} & \textbf{50.38} & \textbf{49.05} & 57.58 & 61.17 & 59.32 \\
\bottomrule
    \end{tabular}
    \caption{Evaluation results compared with baseline methods in terms of precision ($\mathtt{Pre}$, \%), recall ($\mathtt{Rec}$, \%) and F1 score ($\mathtt{F1}$, \%). The best scores are marked in bold. The experimental results of baseline methods, marked with $^*$, are obtained from this work \cite{zhang-etal-2021-aspect-sentiment}.}
    \label{table:results}
\end{table*}

\section{Experiments}

\subsection{Datasets}
We conduct experiments on two public datasets, i.e. $\mathtt{Rest15}$ and $\mathtt{Rest16}$ \cite{zhang-etal-2021-aspect-sentiment}. These two datasets originate from SemEval tasks \cite{pontiki2015semeval,pontiki2016semeval}, which are gradually annotated by previous researchers \cite{peng2020knowing,wan2020target}. After alignment and completion by \citet{zhang-etal-2021-aspect-sentiment}, each instance in the two datasets contains a review sentence, with one or multiple sentiment quadruplets. The statistics are presented in Table \ref{table:data}.

\begin{table}[]
\small
    \centering
    \setlength{\tabcolsep}{1.2mm}{
    \begin{tabular}{l|cccc|cccc}
    \toprule
     & \multicolumn{4}{c|}{$\mathtt{Rest15}$} & \multicolumn{4}{c}{$\mathtt{Rest16}$} \\
    & \#S & \#+ & \#0 & \#- & \#S & \#+ & \#0 & \#- \\
    \midrule
    \textbf{Train} & 834 & 1005 & 34 & 315 & 1264 & 1369 & 62 & 558 \\
    \textbf{Dev} & 209 & 252 & 14 & 81 & 316 & 341 & 23 & 143 \\
    \textbf{Test} & 537 & 453 & 37 & 305 & 544 & 583 & 40 & 176 \\
\bottomrule
    \end{tabular}}
    \caption{Data statistics. \#S, \#+, \#0, and \#- denote the number of sentences, the number of positive, neutral and negative quads respectively.}
    \label{table:data}
\end{table}

\subsection{Implementation Details}
We adopt T5-base \cite{JMLR:v21:20-074} as the pre-trained generative model. The pre-trained parameters are utilized to initialize the model, which is exploited to calculate template orders' entropy without updating any parameters. After selecting order with minimal entropy, we fine-tune T5 with the constructed training samples. The batch size is set to 16. In the pilot experiment, the hyper-parameters are set following \citet{zhang-etal-2021-aspect-sentiment}. The learning rate is set to 3e-4. During the inference, greedy decoding is chosen to generate the output sequence. The number of training epochs is 20 for all experiments. For the proposed approaches, since multiple template orders are combined, we set the learning rate as 1e-4 to prevent overfitting. During the inference, we utilize the beam search decoding, with the number of beam being 5, for generating the output sequence. All reported results are the average of 5 fixed seeds.  


\subsection{Compared Methods}
To make an extensive evaluation, we choose the following strong baseline methods.

\noindent
\begin{itemize}[leftmargin=*]
    \item \textbf{HGCN-BERT+BERT-Linear} \; HGCN \cite{cai2020aspect} aims to jointly extract $ac$ and $sp$. Following it, BERT extracts $at$ and $ot$ \cite{li-etal-2019-exploiting}. Finally, stacking a linear layer (BERT-Linear) forms the full model.
    \item \textbf{HGCN-BERT+BERT-TFM} \; The final stacked layer in the above model is changed to a transformer block (BERT-TFM).
    \item \textbf{TASO-BERT-Linear} \; TAS \cite{wan2020target} is proposed to extract ($ac$, $at$, $sp$) triplets. By changing the tagging schema, it is expanded into TASO (TAS with Opinion). Followed by a linear classification layer, the model is named as TASO-BERT-Linear.
    \item \textbf{TASO-BERT-CRF} \; TASO is followed with a CRF layer, named as TASO-BERT-CRF.
    \item \textbf{Extract-Classify-ACOS} \cite{cai2021aspect} \; It is a two-stage method, which first extracts $at$ and $ot$ from the original sentence. Based on it, $ac$ and $sp$ are obtained through classification. 
    \item \textbf{GAS} \cite{zhang-etal-2021-towards-generative} \; It is the first work to deal aspect level sentiment analysis with generative method, which is modified to directly treat the sentiment quads sequence as the target sequence.
    \item \textbf{Paraphrase} \cite{zhang-etal-2021-aspect-sentiment} \; It is also a generation-based method. By paraphrasing the original sentence, the semantic knowledge from the pre-trained language model can be better exploited.
\end{itemize}




\subsection{Experimental Results}

\begin{table}[]
\small
    \centering
    \setlength{\tabcolsep}{3mm}{
    \begin{tabular}{l|ccc}
    \toprule
    \multirow{2}{*}{Methods} & \multicolumn{3}{c}{$\mathtt{Rest15}$}  \\
    & $\mathtt{Pre}$ & $\mathtt{Rec}$ & $\mathtt{F1}$  \\
    \midrule
    ILO(Entropy Max) & 45.68 & 48.85 & 47.21 \\
    ILO(random) & 46.84 & 50.33 & 48.52 \\
    \midrule
    ILO & \textbf{47.78} & \textbf{50.38} & \textbf{49.05} \\
    \bottomrule
    \toprule
    \multirow{2}{*}{Methods} & \multicolumn{3}{c}{$\mathtt{Rest16}$}  \\
    & $\mathtt{Pre}$ & $\mathtt{Rec}$ & $\mathtt{F1}$  \\
    \midrule
    DLO(Entropy Max) & 57.09 & 60.77 & 58.87 \\
    DLO(random) & 56.92 & 60.91 & 58.85 \\
    \midrule
    DLO & \textbf{57.92} & \textbf{61.80} & \textbf{59.79} \\
\bottomrule
    \end{tabular}}
    \caption{Evaluation results of ablation study.}
    \label{table:results_ablation}
\end{table}

\subsubsection{Overall Results}
Experimental results of various approaches are reported in Table \ref{table:results}. The best scores on each metric are marked in bold. It is worth noting that for our two approaches, i.e. ILO and DLO, the default template orders are selected with the top-3 minimal entropy from all permutations. 

We observe that ILO and DLO achieve the best performances compared with strong baselines. Specifically, comparing with Paraphrase, the absolute improvement of ILO is +2.12\% (+4.51\% relatively) $\mathtt{F1}$ score on $\mathtt{Rest15}$ dataset. ILO outperforms Paraphrase by +1.86\% (+3.21\% relatively) $\mathtt{F1}$ score on $\mathtt{Rest16}$ dataset. This validates the effectiveness of our template-order data augmentation, which provides more informative views for pre-trained models. Our method exploits the order-free property of quadruplet to augment the ``output'' of a model, which is different from the previous data augmentation approaches. 



\begin{figure}[t]
\centering
\subfigure[$\mathtt{Rest15}$]{
\includegraphics[width=0.49\textwidth]{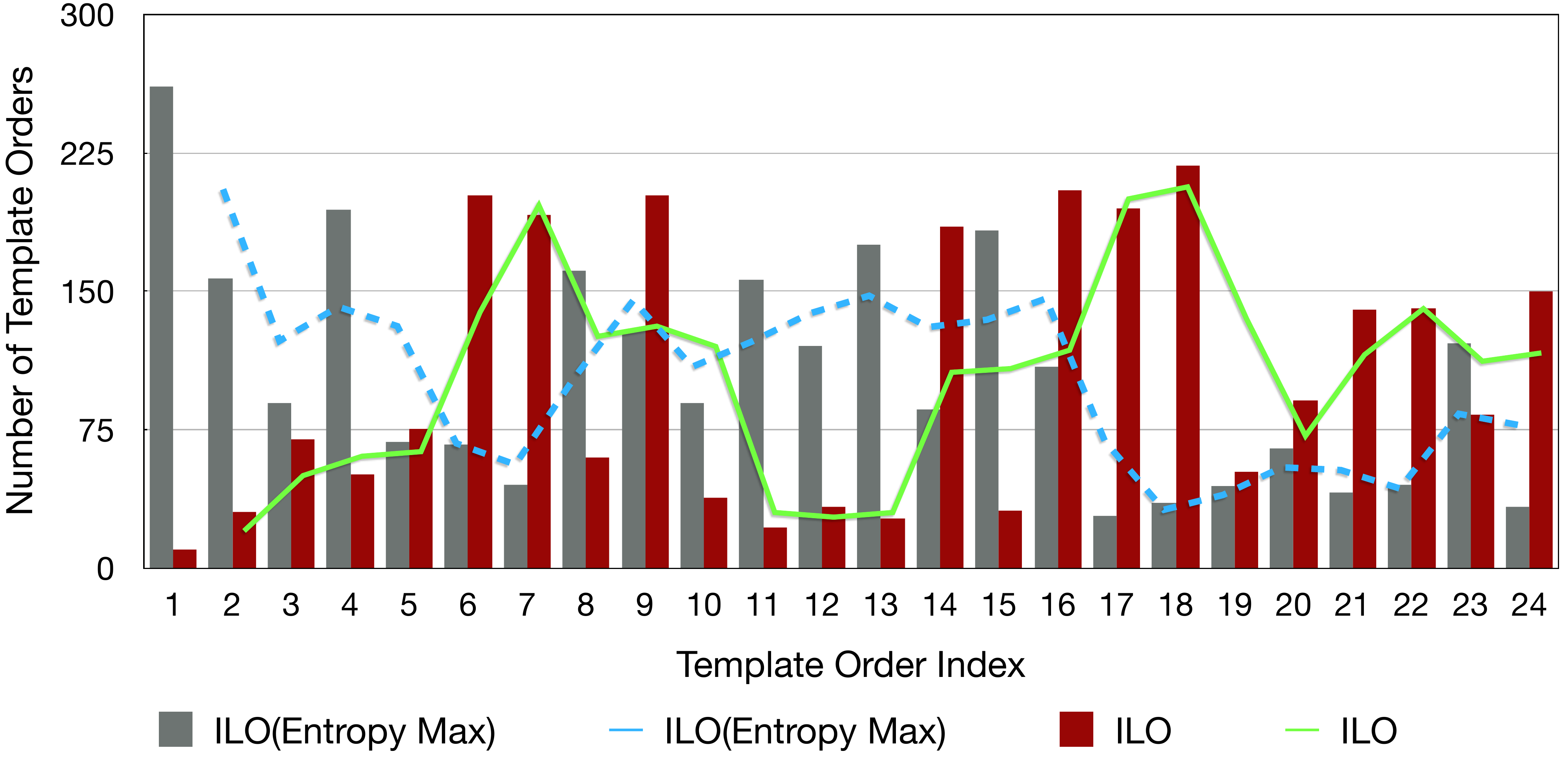}}
\subfigure[$\mathtt{Rest16}$]{
\includegraphics[width=0.49\textwidth]{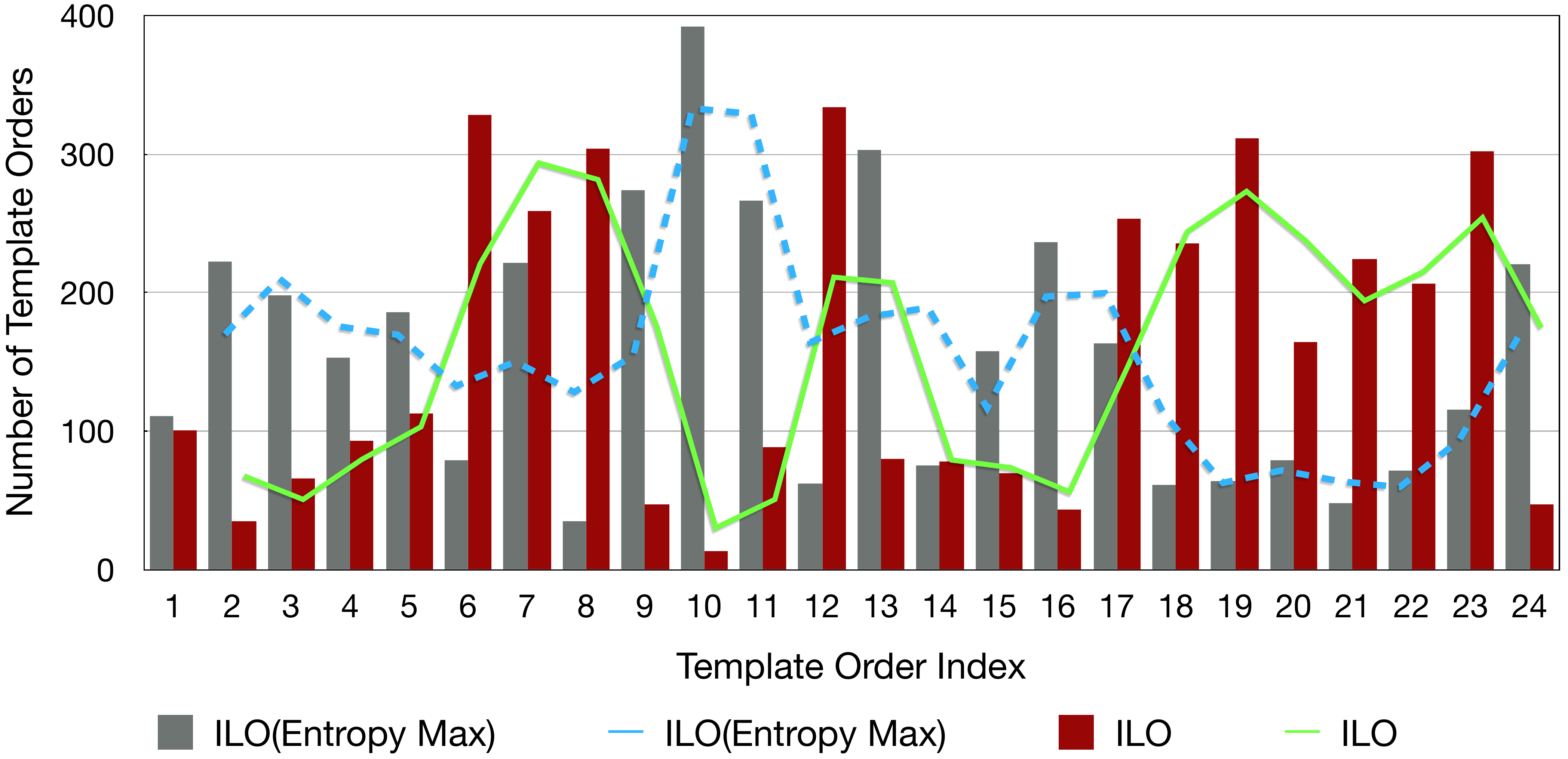}}
\caption{The distribution of the selected templates by ILO and ILO(Entropy Max), respectively, on two datasets. The two curves are the moving average.}
\label{fig:figure_template_scale}
\end{figure}

\subsubsection{Ablation Study}
To further investigate the strategies for selecting template orders, an ablation study is conducted. The results are shown in Table \ref{table:results_ablation}. As aforementioned, the default setting of ILO and DLO is to select top-3  template orders with minimal entropy. The model variants also select top-3 template orders, but with maximal entropy and random sampling. We observe that using minimal entropy consistently outperforms the other two strategies. This verifies that our strategy is effective, and the selected template orders can better promote the potential of T5 on solving the ASQP task.

Moreover, we investigate the distribution of the chosen template orders. Firstly, we sort all the 24 template orders by their $\mathtt{F1}$ scores in ascending order based on the results of the pilot experiment (see Appendix). As depicted in Figure \ref{fig:figure_template_scale}, the horizontal axis represents the template index $i\in{[1,24]}$. The template order at index 1 has the worst performance while index 24 the best. We then count the number of each template index which is selected by ILO. We observe that by minimal entropy, more performant template orders (e.g., index $i\in[17,24]$) are selected compared with using maximal entropy. On the contrary, we also see that ILO chooses less poorly-performed template orders (e.g., index $i\in[1,5]$) than ILO(Entropy Max). This verifies that using minimal entropy, we can select performant template orders. The observations are similar in DLO, which are presented in the appendix.


\begin{table}[]
\small
    \centering
    \setlength{\tabcolsep}{3mm}{
    \begin{tabular}{l|ccc}
    \toprule
    \multirow{2}{*}{Methods} & \multicolumn{3}{c}{$\mathtt{Rest15}$}  \\
    & $\mathtt{Pre}$ & $\mathtt{Rec}$ & $\mathtt{F1}$  \\
    \midrule
    Paraphrase & 35.52 & 37.76 & 36.60 \\
    \midrule
    ILO & 35.66 & 41.05 & 38.16 \\
    ILO(top-10) & \textbf{39.12} & \textbf{43.27} & \textbf{41.08} \\
    \bottomrule
    \toprule
    \multirow{2}{*}{Methods} & \multicolumn{3}{c}{$\mathtt{Rest16}$}  \\
    & $\mathtt{Pre}$ & $\mathtt{Rec}$ & $\mathtt{F1}$  \\
    \midrule
    Paraphrase & 47.87 & 48.96 & 48.40 \\
    \midrule
    ILO & 48.51 & 53.66 & 50.96 \\
    ILO(top-10) & \textbf{51.31} & \textbf{54.29} & \textbf{52.76} \\
\bottomrule
    \end{tabular}}
    \caption{Evaluation results in the low-resource scenario, where we only exploit 25\% of the training data on two datasets.}
    \label{table:results_low}
\end{table}

\subsubsection{Low-Resource Scenario}
\label{sec:low_resource}
To further explore the performances of the proposed method in low-resource settings, we design an experiment which uses only 25\% of the original training data to train the model. The experimental results are shown in Table \ref{table:results_low}. Our approach achieves significant improvements compared with the state-of-the-art. Specifically, ILO(top-10) outperforms Paraphrase by +4.48\% (+12.24\% relatively) and +4.36\% (+9.01\% relatively) $\mathtt{F1}$ scores on $\mathtt{Rest15}$ and $\mathtt{Rest16}$, respectively. This further verifies our hypothesis that different orders provide various informative views of the quadruplet while combining multiple orders as data augmentation can improve the model training, especially in low-resource settings. 

\begin{figure}[t]
\centering
\includegraphics[width=0.48\textwidth]{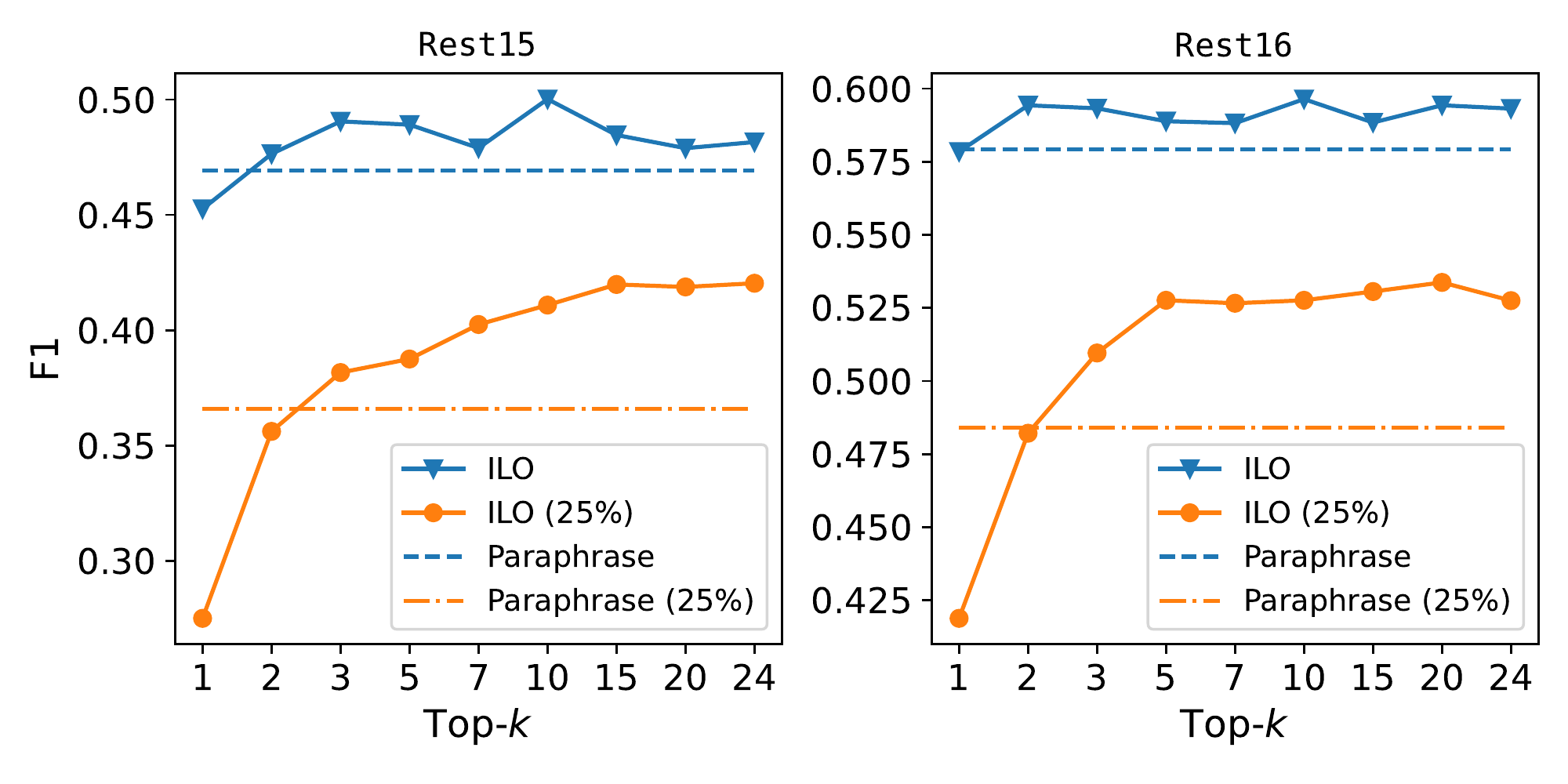} 
\caption{$\mathtt{F1}$ scores of ILO variants by setting different top-$k$. It is worth noting that (25\%) indicates the low-resource scenario, which uses only 25\% training data.}
\label{fig:figure_scale}
\end{figure}

We also plot the $\mathtt{F1}$ score curves by setting different top-$k$ values (see Figure \ref{fig:figure_scale}). It can be seen that under the two settings, i.e. full and 25\% training data, ILO both outperforms Paraphrase. Comparing the two settings, ILO achieves more significant improvements under the low-resource scenario. This observation is in line of expectation. When the training data is adequate, selecting template orders with top-3 is enough. When the training data is limited, model can obtain more gains by setting large $k$. It also shows that our data augmentation is friendly for real applications which have limited labeled data. %

\subsubsection{Effects of Special Marker}
Since we design four special markers for the four elements to jointly train models with multiple templates, we investigate the differences in using other symbols. The templates below are chosen for comparison. \textbf{T2} and \textbf{T3} are inspired by \citet{chia-etal-2022-relationprompt}, which annotate the type of information by specific words.

\begin{itemize}
    \item \textbf{T1}: $\mathtt{[AT]}$ $x_{at}$ $\mathtt{[OT]}$ $x_{ot}$ $\mathtt{[AC]}$ $x_{ac}$ $\mathtt{[SP]}$ $x_{sp}$
    \item \textbf{T2}: aspect term: $x_{at}$ opinion term: $x_{ot}$ aspect category: $x_{ac}$ sentiment polarity: $x_{sp}$
    \item \textbf{T3}: Aspect Term: $x_{at}$ Opinion Term: $x_{ot}$ Aspect Category: $x_{ac}$ Sentiment Polarity: $x_{sp}$
    \item \textbf{T4}: $x_{at}$, $x_{ot}$, $x_{ac}$, $x_{sp}$
\end{itemize}

The evaluation results of the above four templates are reported in Table \ref{table:results_marker}. Firstly, by comparing \textbf{T4} with others, it can be seen that marking the types of four elements are effective for generative ASQP. A possible reason is that marking with either special symbols or specific words helps to demonstrate the structured information \cite{paolini2021structured}. Secondly, \textbf{T1} achieves the best performances on almost all evaluation metrics. Such special markers can avoid overlapping words with sentences. For example, the sentence \emph{``Service is not what one would expect from a joint in this price category.''} contains the word \emph{``category''}, which is overlapped with the type indicator \emph{aspect category} in \textbf{T2}. It is shared between sentence and type markers through word embeddings, and might lead to negative effects.

\begin{table}[]
\small
    \centering
    \setlength{\tabcolsep}{1.5mm}{
    \begin{tabular}{l|ccc|ccc}
    \toprule
    \multirow{2}{*}{Template} & \multicolumn{3}{c|}{$\mathtt{Rest15}$} & \multicolumn{3}{c}{$\mathtt{Rest16}$} \\
    & $\mathtt{Pre}$ & $\mathtt{Rec}$ & $\mathtt{F1}$ & $\mathtt{Pre}$ & $\mathtt{Rec}$ & $\mathtt{F1}$ \\
    \midrule
    \textbf{T1} & \textbf{48.46} & \textbf{49.46} & \textbf{48.95} & \textbf{58.27} & 60.33 & \textbf{59.28} \\
    \midrule
    \textbf{T2} & 47.11 & 48.30 & 47.69 & 57.77 & \textbf{60.40} & 59.05 \\
    \textbf{T3} & 47.62 & 48.60 & 48.10 & 57.09 & 59.57 & 58.20\\
    \textbf{T4} & 46.67 & 47.87 & 47.26 & 57.66 & 59.97 & 58.79 \\

\bottomrule
    \end{tabular}}
    \caption{Evaluation results of special marker.}
    \label{table:results_marker}
\end{table}

\subsubsection{Error Analysis}
\begin{figure}[t]
\centering
\includegraphics[width=0.48\textwidth]{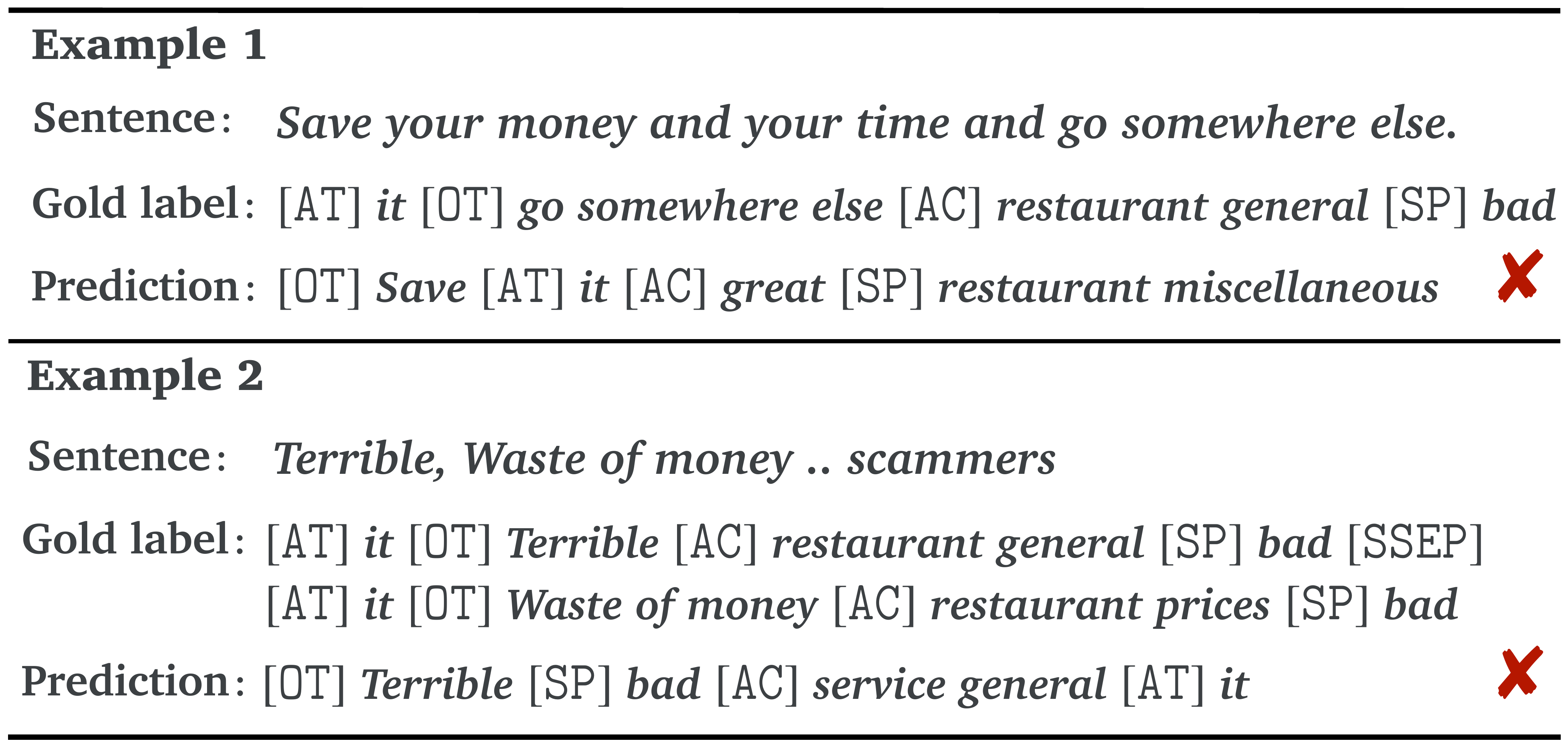} 
\caption{Two error examples predicted by ILO. It is worth noting that the gold label only provides the value of four elements, but does not constrain the order.}
\label{fig:figure_error}
\end{figure}

We further investigate some error cases. Two example cases are presented in Figure \ref{fig:figure_error}. We observe that ILO can generate quadruplets in different orders with the help of special markers. By recognizing the special markers, the quadruplets can be disentangled from the target sequence.

The two examples demonstrate that some cases are still difficult for our approach. The first example contains an implicit aspect term, which is mapped into \emph{``it''}. Its opinion term, i.e. \emph{``go somewhere else''}, also expresses negative sentiment polarity implicitly. This case is wrongly predicted. As for the second one, its gold label consists of two quadruplets. Our method only predicts one quadruplet, which does not match either quadruplet. This example also describes aspect terms implicitly for different aspect categories, i.e. \emph{``restaurant general''} and \emph{``restaurant miscellaneous''}. In summary, sentences with implicit expressions and multiple aspects are usually tough cases. This observation is also consistent with the results from the pilot experiment. As shown in Figure \ref{fig:figure_pre}, the $\mathtt{F1}$ scores of aspect term and opinion term are much worse than other two elements.



\section{Related Work}
\subsection{Aspect-Level Sentiment Analysis}
Aspect-level sentiment analysis presents a research trend that deals with four elements gradually in a finer-grained manner \cite{zhang2022survey}. Analyzing sentiment at the aspect level begins from learning the elements separately \cite{pontiki-etal-2014-semeval}. To name a few, some works have been proposed to classify sentiment polarity given the mentioned aspect, either aspect category \cite{hu-etal-2019-constrained} or aspect term \cite{zhang2020convolution}. Other works extract aspect term \cite{ma-etal-2019-exploring}, classify aspect category \cite{bu-etal-2021-asap}. The four elements are not solely existing, which actually have strong connections with each other. Therefore, researchers focus on learning them jointly, such as aspect sentiment pair \cite{zhao-etal-2020-spanmlt,cai2020aspect} or triplet \cite{chen-qian-2020-relation,mao2021joint}. 

Recently, learning four elements simultaneously sparks new research interests. Two promising directions have been pointed out by researchers. \citet{cai2021aspect} propose a two-stage method by extracting the aspect term and opinion term first. Then these items are utilized to classify aspect category and sentiment polarity. Another method is based on generation model \cite{zhang-etal-2021-aspect-sentiment}. By paraphrasing the input sentence, the quadruplet can be extracted in an end-to-end manner. In this work, we follow the generative direction and consider the order-free property of the quadruplet. To the best of our knowledge, this work is the first to study ASQP from the order perspective.

\subsection{Data Augmentation}
Data augmentation has been widely adopted in both the language and vision fields. We formulate the input and output of a model as $X$ and $Y$, respectively. Previous data augmentation can be divided into three types. The first type is augmenting the input $X$. For example, image flipping, rotation and scaling all change $X$ to seek improvements \cite{shorten2019survey}. In the text tasks, back translation \cite{sugiyama2019data} can also generate pseudo pairs through augmenting $X$. The main idea is that changing $X$ does not affects its ground-truth label $Y$. Secondly, both $X$ and $Y$ are augmented. A promising work is mixup \cite{zhang2018mixup}, which constructs virtual training examples base on the prior knowledge that linear interpolations of feature vectors should lead to linear interpolations of the associated targets. Despite it is intuitive, it has shown effectiveness in many tasks \cite{sun-etal-2020-mixup}. 

The third one is augmenting $Y$. One recent work proposes virtual sequence as the target-side data augmentation \cite{xie2022target} for sequence-to-sequence learning. It deals with typical generation tasks, which are closely connected with the order of words. Different from it, we exploit the characteristic of the generative ASQP task. Order permutations still provide ground-truth labels. Then we think that different orders are just similar to seeing a picture from different perspectives, i.e. different views. Therefore, combining multiple template orders can prevent the model from being biased to superficial patterns, and help it to comprehensively understand the essence of the task. 

\section{Conclusion}
In this work, we study aspect sentiment quad prediction (ASQP) from the template order perspective. We hypothesize that different orders provide various views of the quadruplet. In light of this hypothesis, a simple but effective method is proposed to identify the most proper orders, and further combine multiple proper templates as data augmentation to improve the ASQP task. Specifically, we use the pre-trained language model to select the orders with minimal entropy. By fine-tuning the pre-trained model with these template orders, our model achieves state-of-the-art performances. 

\section*{Limitations}
Our work is the first attempt to improve the ASQP task by combining multiple template orders as data augmentation. Despite state-of-the-art performance, our work still have limitations which may guide the direction of future work.

Firstly, we use the entropy to select the proper template orders. The smaller entropy value indicates that the target sequence is better fitting with the pre-trained language model. However, there may be other criteria for template order selection which can better fine-tune the pre-trained language model to support the ASQP task. 

Secondly, in the experiment, we simply select the top-$k$ template orders for data augmentation. This can be treated as a greedy strategy for the combination. However, each of the top-$k$ orders may not supplement well to each other. More advanced strategies may be designed to select template orders for data augmentation. 


Thirdly, we only consider augmenting the target sequences in the model training, while augmenting both the input and out sequences may bring more performance improvement.  



\section*{Acknowledgements}
We sincerely thank all the anonymous reviewers for providing valuable feedback. This work is supported by the key program of the National Science Fund of Tianjin, China (Grant No. 21JCZDJC00130), the Basic Scientific Research Fund, China (Grant No. 63221028), the National Science and Technology Key Project, China (Grant No. 2021YFB0300104).

\bibliography{anthology,custom}
\bibliographystyle{acl_natbib}

\clearpage
\appendix

\begin{figure}[t]
\centering
\subfigure[$\mathtt{Rest15}$]{
\includegraphics[width=0.49\textwidth]{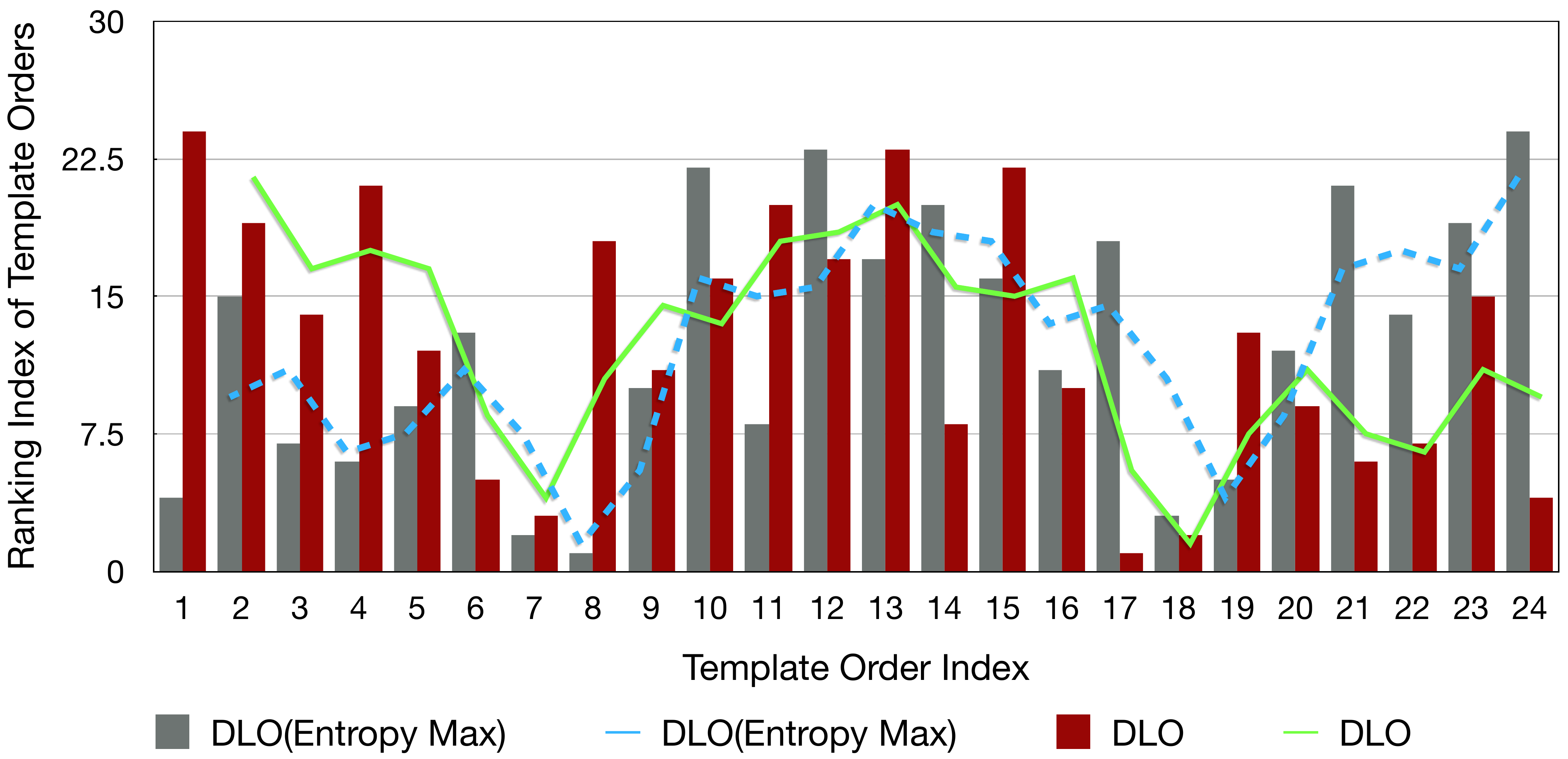}}
\subfigure[$\mathtt{Rest16}$]{
\includegraphics[width=0.49\textwidth]{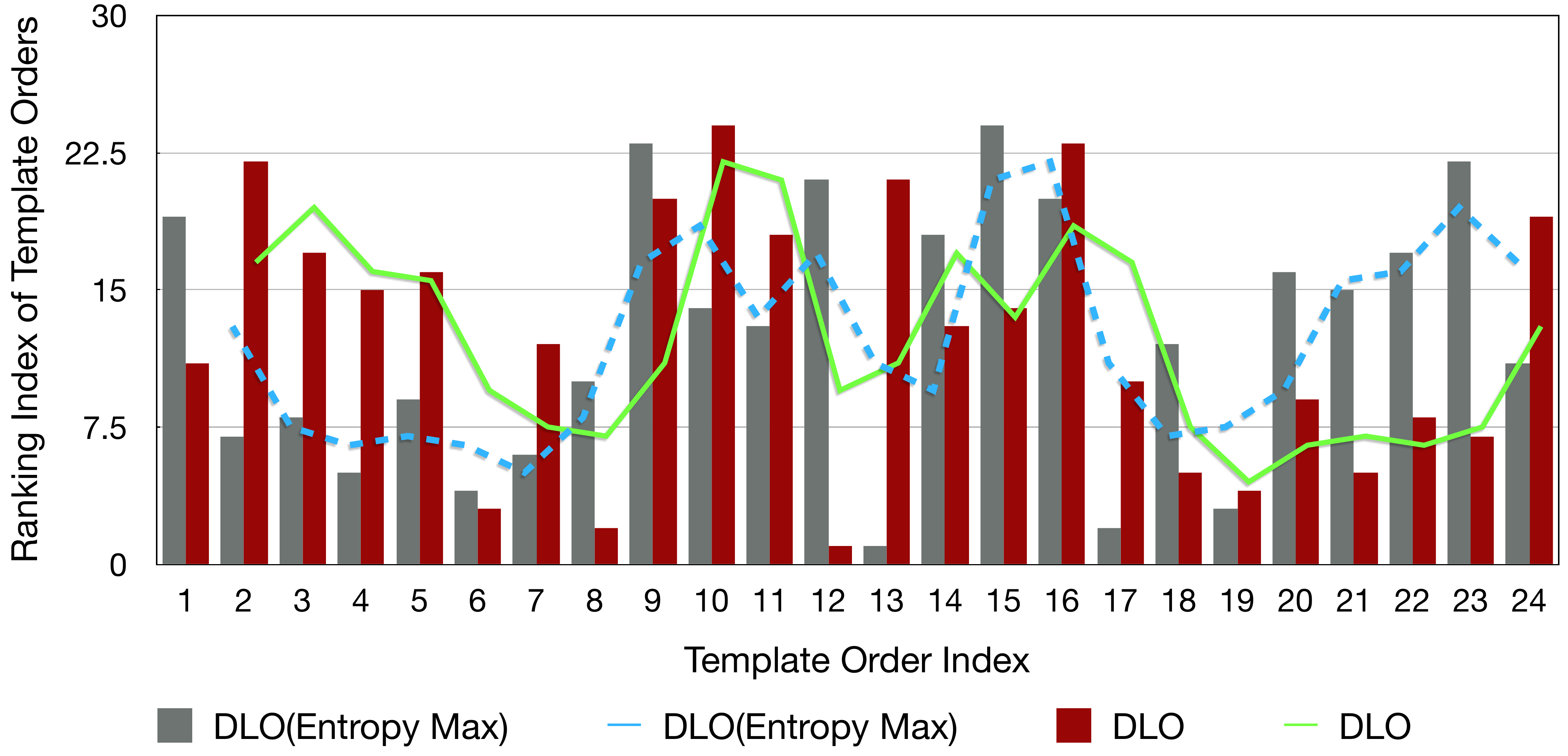}}
\caption{Ranking position of different template orders from DLO and DLO(Entropy Max).}
\label{fig:figure_DLO}
\end{figure}

\section{Appendix}
\subsection{Software and Hardware}
We use Pytorch to implement all the models (Python 3.7). The operating system is Ubuntu 18.04.6. We use a single NVIDIA A6000 GPU with 48GB of RAM.


\subsection{Full Pilot Experimental Results}
The full experimental results of template permutations are presented in Table \ref{table:rest15_appendix_without_marker}, Table \ref{table:rest16_appendix_without_marker}, Table \ref{table:rest15_appendix} and Table \ref{table:rest16_appendix}. The results of each table are sorted by $\mathtt{F1}$ scores in an ascending order.

\subsection{Results of DLO}
Since the proposed DLO choose templates for the whole training set, we plot the ranking position of each template order in Figure \ref{fig:figure_DLO}. Here the horizon axis indicates that the template indexes which are ordered by $\mathtt{F1}$ score from Table \ref{table:rest15_appendix} and Table \ref{table:rest16_appendix} in an ascending order. Then we can see that DLO can choose better-performed templates, where the ranking positions of $O_i, i\in{[17,24]}$ are small. In contrary, DLO(Entropy Max) selects template orders that performed worse.

\begin{table}[]
\small
    \centering
    \begin{tabular}{c|ccc}
    \toprule
    \multirow{2}{*}{Template} & \multicolumn{3}{c}{$\mathtt{Rest15}$} \\ 
    & $\mathtt{Pre}$ & $\mathtt{Rec}$ & $\mathtt{F1}$ \\
    \midrule
$x_{sp}$, $x_{ac}$, $x_{at}$, $x_{ot}$ & 45.55 & 46.34 & 45.94 \\
$x_{ot}$, $x_{at}$, $x_{ac}$, $x_{sp}$ & 45.43 & 47.02 & 46.21 \\ 
$x_{at}$, $x_{ac}$, $x_{sp}$, $x_{ot}$ & 46.15 & 47.02 & 46.58 \\
$x_{ot}$, $x_{ac}$, $x_{at}$, $x_{sp}$ & 46.04 & 47.25 & 46.63 \\
$x_{sp}$, $x_{at}$, $x_{ac}$, $x_{ot}$ & 46.34 & 47.25 & 46.79 \\
$x_{sp}$, $x_{ot}$, $x_{at}$, $x_{ac}$ & 46.12 & 47.52 & 46.81 \\
$x_{ot}$, $x_{ac}$, $x_{sp}$, $x_{at}$ & 46.37 & 47.65 & 47.00 \\
$x_{ot}$, $x_{sp}$, $x_{at}$, $x_{ac}$ & 46.37 & 47.90 & 47.12 \\
$x_{at}$, $x_{sp}$, $x_{ac}$, $x_{ot}$ & 46.58 & 47.67 & 47.12 \\
$x_{sp}$, $x_{at}$, $x_{ot}$, $x_{ac}$ & 46.44 & 47.90 & 47.15 \\
$x_{at}$, $x_{sp}$, $x_{ot}$, $x_{ac}$ & 46.53 & 47.90 & 47.20 \\
$x_{ac}$, $x_{sp}$, $x_{at}$, $x_{ot}$ & 46.67 & 47.85 & 47.25 \\
$x_{at}$, $x_{ot}$, $x_{ac}$, $x_{sp}$ & 46.67 & 47.87 & 47.26 \\
$x_{ot}$, $x_{sp}$, $x_{ac}$, $x_{at}$ & 46.78 & 47.90 & 47.33 \\
$x_{at}$, $x_{ac}$, $x_{ot}$, $x_{sp}$ & 46.72 & 47.97 & 47.34 \\
$x_{ac}$, $x_{at}$, $x_{ot}$, $x_{sp}$ & 46.75 & 48.10 & 47.41 \\
$x_{ac}$, $x_{ot}$, $x_{at}$, $x_{sp}$ & 47.07 & 47.85 & 47.46 \\
$x_{at}$, $x_{ot}$, $x_{sp}$, $x_{ac}$ & 46.91 & 48.23 & 47.56 \\
$x_{ot}$, $x_{at}$, $x_{sp}$, $x_{ac}$ & 46.73 & 48.43 & 47.56 \\
$x_{sp}$, $x_{ac}$, $x_{ot}$, $x_{at}$ & 47.29 & 48.03 & 47.66 \\
$x_{sp}$, $x_{ot}$, $x_{ac}$, $x_{at}$ & 47.29 & 48.20 & 47.74 \\
$x_{ac}$, $x_{at}$, $x_{sp}$, $x_{ot}$ & 47.29 & 48.28 & 47.78 \\
$x_{ac}$, $x_{ot}$, $x_{sp}$, $x_{at}$ & 47.41 & 48.33 & 47.86 \\
$x_{ac}$, $x_{sp}$, $x_{ot}$, $x_{at}$ & 47.60 & 48.75 & 48.17 \\
\bottomrule
    \end{tabular}
    \caption{Evaluation results on $\mathtt{Rest15}$, which are sorted by $\mathtt{F1}$ scores.}
    \label{table:rest15_appendix_without_marker}
\end{table}

\begin{table}[]
\small
    \centering
    \begin{tabular}{c|ccc}
    \toprule
    \multirow{2}{*}{Template} & \multicolumn{3}{c}{$\mathtt{Rest16}$} \\ 
    & $\mathtt{Pre}$ & $\mathtt{Rec}$ & $\mathtt{F1}$ \\
    \midrule
$x_{ot}$, $x_{at}$, $x_{sp}$, $x_{ac}$ & 56.04 & 58.17 & 57.09 \\
$x_{ot}$, $x_{sp}$, $x_{at}$, $x_{ac}$ & 56.15 & 58.52 & 57.31 \\
$x_{at}$, $x_{ac}$, $x_{sp}$, $x_{ot}$ & 56.71 & 58.52 & 57.60 \\
$x_{ac}$, $x_{ot}$, $x_{at}$, $x_{sp}$ & 56.73 & 58.55 & 57.62 \\
$x_{ot}$, $x_{ac}$, $x_{sp}$, $x_{at}$ & 56.78 & 59.00 & 57.87 \\
$x_{ac}$, $x_{ot}$, $x_{sp}$, $x_{at}$ & 57.14 & 58.72 & 57.92 \\
$x_{sp}$, $x_{ot}$, $x_{at}$, $x_{ac}$ & 56.87 & 59.07 & 57.95 \\
$x_{ac}$, $x_{sp}$, $x_{ot}$, $x_{at}$ & 56.89 & 59.10 & 57.98 \\
$x_{ot}$, $x_{at}$, $x_{ac}$, $x_{sp}$ & 56.95 & 59.07 & 57.99 \\
$x_{ac}$, $x_{at}$, $x_{sp}$, $x_{ot}$ & 56.91 & 59.17 & 58.02 \\
$x_{ot}$, $x_{ac}$, $x_{at}$, $x_{sp}$ & 57.14 & 58.97 & 58.04 \\
$x_{sp}$, $x_{ac}$, $x_{at}$, $x_{ot}$ & 57.77 & 59.07 & 58.41 \\
$x_{at}$, $x_{sp}$, $x_{ac}$, $x_{ot}$ & 57.35 & 59.60 & 58.45 \\
$x_{at}$, $x_{ac}$, $x_{ot}$, $x_{sp}$ & 57.49 & 59.57 & 58.51 \\
$x_{ot}$, $x_{sp}$, $x_{ac}$, $x_{at}$ & 57.58 & 59.77 & 58.66 \\
$x_{sp}$, $x_{at}$, $x_{ac}$, $x_{ot}$ & 57.70 & 59.87 & 58.76 \\
$x_{sp}$, $x_{ac}$, $x_{ot}$, $x_{at}$ & 57.71 & 59.87 & 58.77 \\
$x_{at}$, $x_{ot}$, $x_{ac}$, $x_{sp}$ & 57.67 & 59.98 & 58.80 \\
$x_{sp}$, $x_{ot}$, $x_{ac}$, $x_{at}$ & 58.14 & 59.82 & 58.97 \\
$x_{at}$, $x_{ot}$, $x_{sp}$, $x_{ac}$ & 57.68 & 60.35 & 58.99 \\
$x_{sp}$, $x_{at}$, $x_{ot}$, $x_{ac}$ & 57.93 & 60.13 & 59.01 \\
$x_{ac}$, $x_{at}$, $x_{ot}$, $x_{sp}$ & 58.12 & 60.08 & 59.08 \\
$x_{at}$, $x_{sp}$, $x_{ot}$, $x_{ac}$ & 57.95 & 60.33 & 59.11 \\
$x_{ac}$, $x_{sp}$, $x_{at}$, $x_{ot}$ & 58.11 & 60.33 & 59.20 \\
\bottomrule
    \end{tabular}
    \caption{Evaluation results on $\mathtt{Rest16}$, which are sorted by $\mathtt{F1}$ scores.}
    \label{table:rest16_appendix_without_marker}
\end{table}

\begin{table*}[]
\small
    \centering
    \begin{tabular}{c|ccc}
    \toprule
    \multirow{2}{*}{Template} & \multicolumn{3}{c}{$\mathtt{Rest15}$} \\ 
    & $\mathtt{Pre}$ & $\mathtt{Rec}$ & $\mathtt{F1}$ \\
    \midrule
$\mathtt{[SP]}$ $x_{sp}$ $\mathtt{[AT]}$ $x_{at}$ $\mathtt{[AC]}$ $x_{ac}$ $\mathtt{[OT]}$ $x_{ot}$ & 45.34 & 46.39 & 45.86 \\
$\mathtt{[SP]}$ $x_{sp}$ $\mathtt{[AC]}$ $x_{ac}$ $\mathtt{[OT]}$ $x_{ot}$ $\mathtt{[AT]}$ $x_{at}$ & 46.24 & 46.67 & 46.45 \\
$\mathtt{[AC]}$ $x_{ac}$ $\mathtt{[SP]}$ $x_{sp}$ $\mathtt{[OT]}$ $x_{ot}$ $\mathtt{[AT]}$ $x_{at}$ & 46.28 & 47.37 & 46.81 \\
$\mathtt{[SP]}$ $x_{sp}$ $\mathtt{[OT]}$ $x_{ot}$ $\mathtt{[AT]}$ $x_{at}$ $\mathtt{[AC]}$ $x_{ac}$ & 46.49 & 47.47 & 46.98 \\
$\mathtt{[AT]}$ $x_{at}$ $\mathtt{[SP]}$ $x_{sp}$ $\mathtt{[AC]}$ $x_{ac}$ $\mathtt{[OT]}$ $x_{ot}$ & 46.40 & 47.85 & 47.11 \\
$\mathtt{[OT]}$ $x_{ot}$ $\mathtt{[SP]}$ $x_{sp}$ $\mathtt{[AC]}$ $x_{ac}$ $\mathtt{[AT]}$ $x_{at}$ & 46.64 & 47.62 & 47.12 \\
$\mathtt{[OT]}$ $x_{ot}$ $\mathtt{[AT]}$ $x_{at}$ $\mathtt{[SP]}$ $x_{sp}$ $\mathtt{[AC]}$ $x_{ac}$ & 46.97 & 47.75 & 47.35 \\
$\mathtt{[AC]}$ $x_{ac}$ $\mathtt{[AT]}$ $x_{at}$ $\mathtt{[SP]}$ $x_{sp}$ $\mathtt{[OT]}$ $x_{ot}$ & 46.90 & 47.97 & 47.43 \\
$\mathtt{[AT]}$ $x_{at}$ $\mathtt{[AC]}$ $x_{ac}$ $\mathtt{[SP]}$ $x_{sp}$ $\mathtt{[OT]}$ $x_{ot}$ & 47.00 & 48.08 & 47.53 \\
$\mathtt{[AC]}$ $x_{ac}$ $\mathtt{[OT]}$ $x_{ot}$ $\mathtt{[SP]}$ $x_{sp}$ $\mathtt{[AT]}$ $x_{at}$ & 47.14 & 48.23 & 47.67 \\
$\mathtt{[SP]}$ $x_{sp}$ $\mathtt{[AT]}$ $x_{at}$ $\mathtt{[OT]}$ $x_{ot}$ $\mathtt{[AC]}$ $x_{ac}$ & 47.06 & 48.40 & 47.72 \\
$\mathtt{[AC]}$ $x_{ac}$ $\mathtt{[SP]}$ $x_{sp}$ $\mathtt{[AT]}$ $x_{at}$ $\mathtt{[OT]}$ $x_{ot}$ & 47.13 & 48.40 & 47.76 \\
$\mathtt{[SP]}$ $x_{sp}$ $\mathtt{[AC]}$ $x_{ac}$ $\mathtt{[AT]}$ $x_{at}$ $\mathtt{[OT]}$ $x_{ot}$ & 47.27 & 48.35 & 47.80 \\
$\mathtt{[OT]}$ $x_{ot}$ $\mathtt{[SP]}$ $x_{sp}$ $\mathtt{[AT]}$ $x_{at}$ $\mathtt{[AC]}$ $x_{ac}$ & 47.51 & 48.18 & 47.84 \\
$\mathtt{[SP]}$ $x_{sp}$ $\mathtt{[OT]}$ $x_{ot}$ $\mathtt{[AC]}$ $x_{ac}$ $\mathtt{[AT]}$ $x_{at}$ & 47.53 & 48.18 & 47.85 \\
$\mathtt{[AT]}$ $x_{at}$ $\mathtt{[AC]}$ $x_{ac}$ $\mathtt{[OT]}$ $x_{ot}$ $\mathtt{[SP]}$ $x_{sp}$ & 47.77 & 48.30 & 48.03 \\
$\mathtt{[OT]}$ $x_{ot}$ $\mathtt{[AT]}$ $x_{at}$ $\mathtt{[AC]}$ $x_{ac}$ $\mathtt{[SP]}$ $x_{sp}$ & 47.57 & 48.55 & 48.06 \\
$\mathtt{[AT]}$ $x_{at}$ $\mathtt{[OT]}$ $x_{ot}$ $\mathtt{[SP]}$ $x_{sp}$ $\mathtt{[AC]}$ $x_{ac}$ & 47.46 & 48.75 & 48.10 \\
$\mathtt{[AC]}$ $x_{ac}$ $\mathtt{[OT]}$ $x_{ot}$ $\mathtt{[AT]}$ $x_{at}$ $\mathtt{[SP]}$ $x_{sp}$ & 48.14 & 48.20 & 48.17 \\
$\mathtt{[OT]}$ $x_{ot}$ $\mathtt{[AC]}$ $x_{ac}$ $\mathtt{[SP]}$ $x_{sp}$ $\mathtt{[AT]}$ $x_{at}$ & 47.93 & 48.75 & 48.33 \\
$\mathtt{[AT]}$ $x_{at}$ $\mathtt{[SP]}$ $x_{sp}$ $\mathtt{[OT]}$ $x_{ot}$ $\mathtt{[AC]}$ $x_{ac}$ & 48.36 & 49.08 & 48.72 \\
$\mathtt{[OT]}$ $x_{ot}$ $\mathtt{[AC]}$ $x_{ac}$ $\mathtt{[AT]}$ $x_{at}$ $\mathtt{[SP]}$ $x_{sp}$ & 48.55 & 48.91 & 48.73 \\
$\mathtt{[AC]}$ $x_{ac}$ $\mathtt{[AT]}$ $x_{at}$ $\mathtt{[OT]}$ $x_{ot}$ $\mathtt{[SP]}$ $x_{sp}$ & 48.58 & 49.01 & 48.79 \\
$\mathtt{[AT]}$ $x_{at}$ $\mathtt{[OT]}$ $x_{ot}$ $\mathtt{[AC]}$ $x_{ac}$ $\mathtt{[SP]}$ $x_{sp}$ & 48.46 & 49.46 & 48.95 \\
\bottomrule
    \end{tabular}
    \caption{Evaluation results on $\mathtt{Rest15}$, which are sorted by $\mathtt{F1}$ scores.}
    \label{table:rest15_appendix}
\end{table*}

\begin{table*}[]
\small
    \centering
    \begin{tabular}{c|ccc}
    \toprule
    \multirow{2}{*}{Template} & \multicolumn{3}{c}{$\mathtt{Rest16}$} \\ 
    & $\mathtt{Pre}$ & $\mathtt{Rec}$ & $\mathtt{F1}$ \\
    \midrule
$\mathtt{[AT]}$ $x_{at}$ $\mathtt{[SP]}$ $x_{sp}$ $\mathtt{[AC]}$ $x_{ac}$ $\mathtt{[OT]}$ $x_{ot}$ & 56.36 & 58.80 & 57.55 \\
$\mathtt{[SP]}$ $x_{sp}$ $\mathtt{[AC]}$ $x_{ac}$ $\mathtt{[OT]}$ $x_{ot}$ $\mathtt{[AT]}$ $x_{at}$ & 57.17 & 58.97 & 58.06 \\
$\mathtt{[AC]}$ $x_{ac}$ $\mathtt{[SP]}$ $x_{sp}$ $\mathtt{[AT]}$ $x_{at}$ $\mathtt{[OT]}$ $x_{ot}$ & 57.39 & 58.80 & 58.08 \\
$\mathtt{[AC]}$ $x_{ac}$ $\mathtt{[SP]}$ $x_{sp}$ $\mathtt{[OT]}$ $x_{ot}$ $\mathtt{[AT]}$ $x_{at}$ & 57.07 & 59.17 & 58.10 \\
$\mathtt{[AC]}$ $x_{ac}$ $\mathtt{[AT]}$ $x_{at}$ $\mathtt{[OT]}$ $x_{ot}$ $\mathtt{[SP]}$ $x_{sp}$ & 57.20 & 59.32 & 58.24 \\
$\mathtt{[OT]}$ $x_{ot}$ $\mathtt{[SP]}$ $x_{sp}$ $\mathtt{[AC]}$ $x_{ac}$ $\mathtt{[AT]}$ $x_{at}$ & 57.31 & 59.25 & 58.26 \\
$\mathtt{[AT]}$ $x_{at}$ $\mathtt{[AC]}$ $x_{ac}$ $\mathtt{[SP]}$ $x_{sp}$ $\mathtt{[OT]}$ $x_{ot}$ & 57.30 & 59.35 & 58.31 \\
$\mathtt{[OT]}$ $x_{ot}$ $\mathtt{[AT]}$ $x_{at}$ $\mathtt{[AC]}$ $x_{ac}$ $\mathtt{[SP]}$ $x_{sp}$ & 57.29 & 59.42 & 58.34 \\
$\mathtt{[SP]}$ $x_{sp}$ $\mathtt{[OT]}$ $x_{ot}$ $\mathtt{[AC]}$ $x_{ac}$ $\mathtt{[AT]}$ $x_{at}$ & 57.40 & 59.52 & 58.44 \\
$\mathtt{[SP]}$ $x_{sp}$ $\mathtt{[AT]}$ $x_{at}$ $\mathtt{[AC]}$ $x_{ac}$ $\mathtt{[OT]}$ $x_{ot}$ & 57.77 & 59.17 & 58.46 \\
$\mathtt{[OT]}$ $x_{ot}$ $\mathtt{[AT]}$ $x_{at}$ $\mathtt{[SP]}$ $x_{sp}$ $\mathtt{[AC]}$ $x_{ac}$ & 57.63 & 59.82 & 58.70 \\
$\mathtt{[AC]}$ $x_{ac}$ $\mathtt{[AT]}$ $x_{at}$ $\mathtt{[SP]}$ $x_{sp}$ $\mathtt{[OT]}$ $x_{ot}$ & 57.64 & 59.80 & 58.70 \\
$\mathtt{[SP]}$ $x_{sp}$ $\mathtt{[OT]}$ $x_{ot}$ $\mathtt{[AT]}$ $x_{at}$ $\mathtt{[AC]}$ $x_{ac}$ & 57.81 & 59.95 & 58.86 \\
$\mathtt{[AC]}$ $x_{ac}$ $\mathtt{[OT]}$ $x_{ot}$ $\mathtt{[AT]}$ $x_{at}$ $\mathtt{[SP]}$ $x_{sp}$ & 58.04 & 59.82 & 58.92 \\
$\mathtt{[AC]}$ $x_{ac}$ $\mathtt{[OT]}$ $x_{ot}$ $\mathtt{[SP]}$ $x_{sp}$ $\mathtt{[AT]}$ $x_{at}$ & 57.87 & 60.02 & 58.93 \\
$\mathtt{[SP]}$ $x_{sp}$ $\mathtt{[AC]}$ $x_{ac}$ $\mathtt{[AT]}$ $x_{at}$ $\mathtt{[OT]}$ $x_{ot}$ & 58.29 & 59.80 & 59.03 \\
$\mathtt{[AT]}$ $x_{at}$ $\mathtt{[AC]}$ $x_{ac}$ $\mathtt{[OT]}$ $x_{ot}$ $\mathtt{[SP]}$ $x_{sp}$ & 58.11 & 60.02 & 59.05 \\
$\mathtt{[OT]}$ $x_{ot}$ $\mathtt{[AC]}$ $x_{ac}$ $\mathtt{[AT]}$ $x_{at}$ $\mathtt{[SP]}$ $x_{sp}$ & 58.13 & 60.10 & 59.10 \\
$\mathtt{[AT]}$ $x_{at}$ $\mathtt{[OT]}$ $x_{ot}$ $\mathtt{[SP]}$ $x_{sp}$ $\mathtt{[AC]}$ $x_{ac}$ & 58.25 & 60.05 & 59.14 \\
$\mathtt{[OT]}$ $x_{ot}$ $\mathtt{[AC]}$ $x_{ac}$ $\mathtt{[SP]}$ $x_{sp}$ $\mathtt{[AT]}$ $x_{at}$ & 57.64 & 60.85 & 59.20 \\
$\mathtt{[AT]}$ $x_{at}$ $\mathtt{[OT]}$ $x_{ot}$ $\mathtt{[AC]}$ $x_{ac}$ $\mathtt{[SP]}$ $x_{sp}$ & 58.27 & 60.33 & 59.28 \\
$\mathtt{[AT]}$ $x_{at}$ $\mathtt{[SP]}$ $x_{sp}$ $\mathtt{[OT]}$ $x_{ot}$ $\mathtt{[AC]}$ $x_{ac}$ & 58.03 & 60.63 & 59.30 \\
$\mathtt{[OT]}$ $x_{ot}$ $\mathtt{[SP]}$ $x_{sp}$ $\mathtt{[AT]}$ $x_{at}$ $\mathtt{[AC]}$ $x_{ac}$ & 58.72 & 60.45 & 59.57 \\
$\mathtt{[SP]}$ $x_{sp}$ $\mathtt{[AT]}$ $x_{at}$ $\mathtt{[OT]}$ $x_{ot}$ $\mathtt{[AC]}$ $x_{ac}$ & 58.50 & 60.75 & 59.60 \\
\bottomrule
    \end{tabular}
    \caption{Evaluation results on $\mathtt{Rest16}$, which are sorted by $\mathtt{F1}$ scores.}
    \label{table:rest16_appendix}
\end{table*}

\end{document}